\documentclass[letterpaper, 10 pt, conference]{ieeeconf}  

\IEEEoverridecommandlockouts                              

\usepackage[table,xcdraw]{xcolor}
\usepackage{multirow}
\usepackage{booktabs}
\usepackage{bm}
\usepackage{amsmath}
\usepackage{censor}
\usepackage{amssymb}
\usepackage{graphicx}
\usepackage{array}
\usepackage{mathtools}
\usepackage{hyperref}
\newcounter{subfigref}[figure]

\hypersetup{
    colorlinks=true,  
    linkcolor=black,  
    citecolor=black,  
    filecolor=black,  
    urlcolor=blue     
}
\usepackage{float}
\usepackage{arydshln}

\title{\LARGE \bf
What do VLM-Based Vision-Language Navigation Models Rely on: Interpreting and Steering Policy Behavior
}

\StopCensoring 
\author{
\censor{
Débora Oliveira Makowski*$^{1}$, Samiran Gode*$^{1}$, Abhijeet Nayak*$^{1}$, \\ Marco Hutter$^{2}$, Cordelia Schmid$^{3}$, Lukas Rosenberger-Schmid$^{1}$ and Wolfram Burgard$^{1}$
}
\thanks{\censor{${1}$ University of Technology Nuremberg.}}%
\thanks{\censor{${2}$ ETH Zürich.} }%
\thanks{\censor{${3}$ Inria, Ecole Normale Supérieure, CNRS, PSL Research University. }}%
\thanks{\censor{* These authors contributed equally to the work. Listing order is random.}}%
\thanks{\censor{\url{https://utn-air.github.io/vln-interpretability}}}%
}

\begin{document}

\maketitle
\thispagestyle{empty}
\pagestyle{empty}

\begin{abstract}
Modern Vision-Language Navigation~(VLN) models rely mostly on pre-trained large Vision-Language Models~(VLMs) to predict navigation actions.
While this fusion of language instructions and visual observations allows multimodal reasoning, it obscures how information is routed across modalities or what mechanisms drive navigation decisions. 
Thus, it remains unclear whether VLN models ground their predictions in relevant semantic cues or can track task progress. 
In this work, we study the interpretability and steerability of VLN models.
We use intervention-based metrics that measure how visual observations, instructions, and visual memory causally influence navigation decisions.
Our results show that these navigation policies are sensitive to all input modalities and do not depend on a single one.
We further show that these agents encode navigation progress and retain semantic structure from their VLM backbones, enabling concept-level steering through internal activations. 
Finally, we extract activation vectors for abstract behaviors to transfer them zero-shot to out-of-distribution real-world scenarios, improving performance without additional fine-tuning.

\end{abstract}
\section{Introduction}

Vision-Language-Navigation (VLN)~\cite{zheng2025efficientvln, chu2026abot} models have demonstrated increased embodied intelligence on navigation benchmarks~\cite{R2R,ku2020rxr}.
This progress has largely been driven by pretrained large-scale Vision-Language Models (VLMs) that jointly process natural-language instruction and visual observations to predict navigation actions end-to-end.
Modern VLN architectures further introduce reasoning~\cite{wu2026dual, guo2026awarevln, huang2026tic}, linguistic~\cite{wang2026vlingnav} and spatial memories~\cite{zeng2026janusvln} to encourage semantic grounding, task decomposition, and progress tracking.

While these benchmarks primarily evaluate models through closed-loop success rate, improvements in navigation performance alone do not prove that these components function as intended. 
A model may rely on relevant landmarks, exploit correlations learned during training, or compensate for missing information through another modality, and still produce the same trajectory. 
Understanding what the model attends to and relies on internally is necessary to better characterize the mechanisms underlying embodied reasoning.

Yet, these mechanisms remain poorly understood: it is unclear how VLN models route information between modalities, how navigation concepts are stored, and whether these representations causally influence action predictions.
Analyzing these questions can reveal how VLN models internally process information during navigation, complementing usual benchmark evaluation.
Moreover, navigation failures may not necessarily reflect the absence of task-relevant concepts; instead, a model may encode the required information but fail to use it effectively during action selection.
In such cases, steering internal representations could provide an alternative to collecting additional data to retrain the model.

\begin{figure}[t]
    \centering
    \vspace{5.1pt}
    \includegraphics[width=.95\linewidth]{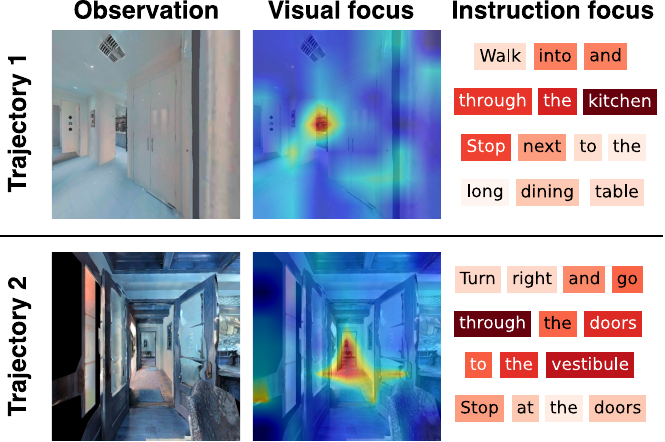}
    \caption{Attribution heatmaps generated with ISS~\cite{zhang2026iss} for CorrectNav~\cite{yu2026correctnav}. 
    Red indicates higher sensitivity and blue lower. 
    In text, darker tokens indicate higher sensitivity. 
    Active subtask verbs (e.g., \textit{stop}, \textit{through}) and semantic cues (e.g., \textit{kitchen}, \textit{doors}, \textit{vestibule}) have higher sensitivity. 
    Note that even distant objects (\textit{kitchen}, in traj. 1) activate their textual concepts. }
    \label{fig:iss_text}
    \vspace{-1em}

\end{figure}
To understand the internal mechanisms of VLN policies, we conduct a study that combines input attribution and causal interventions to quantify how visual observations, instructions, and memory influence actions. 
We steer the model using semantic concepts and hidden state activations within the model.
Our main contributions are:
\begin{itemize}
    \item The first causal interpretability study of VLM-based VLNs on the sensitivity and robustness to changes in the model input, specifically semantic cues.  
    \item A steering study that investigates navigation-relevant representations that persist after VLM finetuning and how these representations can be used to influence model behavior.
    \item An activation steering-based real-world deployment that induces navigation recovery without updating model parameters in out-of-domain settings.
\end{itemize} 

\section{Related Work}

\subsection{Vision-and-Language Navigation}
\label{sec:related_vln}

Vision-Language Navigation (VLN) requires an embodied agent to navigate through an environment by following a natural-language instruction~\cite{R2R,gu2022vision}. 
Recent approaches build VLN agents around Vision-Language Models (VLMs). 
NaVid~\cite{zhang2024navid} introduces a video-based VLM that predicts navigation actions. 
NaVILA~\cite{cheng2024navila} extends VLA-based navigation to legged robots. 
CorrectNav~\cite{yu2026correctnav} learns from erroneous trajectories, while AwareVLN~\cite{guo2026awarevln} introduces reasoning behavior. 
DualVLN~\cite{wei2026ground} uses a VLM along with a diffusion-based action expert, whereas Efficient-VLN~\cite{zheng2025efficientvln} focuses on reducing the computational cost of processing long navigation histories. 
Together, these developments demonstrate the increasing capability of VLM-based navigation agents.

\subsection{Interpretability and Grounding Analysis}
\label{sec:related_interpretability}

Several interpretability techniques have been proposed for understanding deep neural networks.
Previous works~\cite{sundararajan2017ig,selvaraju2017grad} compute input relevance by using gradient-based methods.
Perturbation-based methods~\cite{petsiuk2018rise} estimate feature importance by observing changes under randomized masking.
More recent methods~\cite{li2025tam,chen2026eagle} extend interpretability to autoregressive models.
However, these methods were primarily developed and evaluated on prediction tasks such as image classification, rather than embodied tasks.

Few works analyze the internal grounding of VLN agents. 
Hu~\textit{et~al.}~\cite{hu2019you} study how language grounds to visual appearance and route structure. 
Zhu~\textit{et~al.}~\cite{zhu2022diagnosing} identify imbalances between visual and textual attention, reinforcing the open question about whether attention maps alone provide reliable evidence of explanation~\cite{jain2019attention}.
Chen~\textit{et~al.}~\cite{chen2023evaluating} evaluate explanation methods for transformer-based VLN.
Nevertheless, none of these methods evaluate VLM-based VLNs.
Recent work~\cite{zhang2026iss} has studied interpretability in Vision-Language-Action (VLA) models. 
Their experiments focus on robotic manipulation tasks rather than VLN, but provide an interventional approach for studying whether embodied policies rely on task-relevant visual evidence.

\subsection{Steering}
Steering is a popular method to influence model behavior.
For LLMs, previous work~\cite{subramani-etal-2022-extracting,LiInference-TimeInterventionElicitingTruthful} extracted latent steering directions and applied inference-time interventions, while others~\cite{ActAddturner2023steering,zou2023representation,CAArimsky-etal-2024-steering} found contrastive vectors for activation-based steering. 
Subsequent work identified low-dimensional directions~\cite{RefusalActivationArditi} while also exposing limitations in the generalization and reliability of steering vectors~\cite{TanReliabilitySteeringVectors}.
Recent work~\cite{haon2025mechanistic} explores steering for VLAs by using value vectors from the VLM backbone to control action prediction. 
Other work~\cite{buurmeijer2026observing} uses activation steering to propose a linear observer and controller for latent states. 
Hong~\textit{et~al.}~\cite{LQRhong2026steeringWAM} model transformer dynamics and use LQR to steer World Action Models. 
Despite steady progress in steering methods for VLAs, steering for VLNs remains underexplored.

\section{Overview}

At each navigation step $t$, a VLN policy $\pi_\theta$ receives a natural language instruction $I$ together with a selected set of previously observed images.
Let $\mathcal{T}_I$ denote the tokenized representation of the instruction $I$.
The agent has observed a sequence of images $\mathcal{O}_0,\ldots,\mathcal{O}_{t}$.
Each VLN pipeline uses a memory-sampling strategy to select a uniformly sampled subset of the visual history, $\mathcal{S}_t \subseteq \{0,\ldots,t\}$.
The corresponding visual memory is
\begin{equation}
    \mathcal{M}_t=\left\{\mathcal{O}_\tau\mid\tau\in\mathcal{S}_t\right\}.
\end{equation}
The vision encoder processes each observation in $\mathcal{M}_t$ to generate vision tokens $\mathcal{T}_{\mathcal{M}_t}$.
The instruction and vision tokens are then arranged according to the prompt format of the particular VLN pipeline as
\begin{equation}
    C_t = \text{Prompt}_{\theta} \left( \mathcal{T}_I, \mathcal{T}_{\mathcal{M}_t} \right),
\end{equation}
where $C_t$ denotes the complete multimodal context provided to the VLM at step $t$, and $\text{Prompt}_\theta$ represents pipeline-specific choices such as ordering, formatting, and serialization of the instruction and visual tokens.
Conditioned on $C_t$, the model autoregressively generates an output sequence 
\begin{equation}
    \mathcal{T}_\text{out} \sim \pi_\theta(\cdot\mid C_t) = \left(y_1, \ldots, y_{n_t} \right),
\end{equation}
where $n_t$ is the number of output tokens.
The action sequence $\mathcal{A}_{t:t+h-1}$ is decoded from the output tokens, where $h$ denotes the action horizon and is specific to $\pi_\theta$.
The action sequence is a set of discrete actions $\in$ \{\textit{forward}, \textit{left}, \textit{right}, \textit{stop}\}.

\section{Methodology}

\subsection{Interpretability}

To investigate which input tokens VLNs focus on, we use three attribution methods:

\subsubsection{Interventional Significance Score (ISS)~\cite{zhang2026iss}}
We perform a set of $K$ interventions on the context $C_t$.
For each intervention $k\in\{1,\ldots,K\}$, a group of input tokens $\mathcal{G}_k\subseteq C_t$ are selected to define an intervention mask $M_k$.
The perturbed context is denoted as $\widetilde{C}_t^{\,k}=\Phi(C_t,M_k)$, where $\Phi$ is the modality-specific perturbation operator.
Visual tokens may be blurred, while text tokens may be deleted or replaced with antonyms.
Applying the policy $\pi_\theta$ to $\widetilde{C}_t^{\,k}$ produces the perturbed action sequence $\widetilde{\mathcal{A}}_{t:t+h-1}^{\,k}$.

The effect of intervention $k$ is quantified using a discrepancy function that measures the sensitivity of the predicted trajectory to the input component corrupted by $\Phi$.
Consider $a_{t+j}\in\mathbb{R}^{2}$ denotes the end-point after executing $\mathcal{A}_{t:t+j}$ discrete actions, while $\widetilde{a}_{t+j}^{k}$ denotes the corresponding end-point under intervention $k$.
We define the intervention score as the accumulated positional and angular divergence between the original and perturbed trajectories
\begin{equation}
\delta_k = \frac{1}{h}\sum_{j=0}^{h-1} \Big[ \operatorname{MSE}\left(a_{t+j},\widetilde{a}_{t+j}^{\,k}\right) + \lambda\,\measuredangle\big(\phi_{t+j},\widetilde{\phi}_{t+j}^{\,k}\big) \Big], 
\end{equation}
with $\lambda$ being the ratio between a forward step and a turning unit step of policy $\theta$, $\phi$ the agent's angle, and $\measuredangle$ is the wrapped angular difference.
The final score for token $i$ is obtained by averaging over $K$ sampled masks where $i\in\mathcal{G}_k$.

\subsubsection{Integrated Gradients (IG)~\cite{sundararajan2017ig}}
IG computes attributions by integrating the gradients of the model output along a path from a corrupted observation to the original observation.
We define the embeddings of the tokens in context $C_t$ as $ E_t\in\mathbb{R}^{n_t\times d} $. 
Similarly, we denote $ \widetilde{E}_t$
as their corrupted embeddings. 
We construct a linear interpolation as
\begin{equation}
\widehat{E}_t^{\,n} = \widetilde{E}_t + \frac{n}{N}(E_t-\widetilde{E}_t),
\end{equation}
where $N$ is the number of steps and $n\in\{0,\ldots,N\}$. 
For each interpolated observation ${\widehat{E}}_{t}^{n}$, we perform a forward pass through the policy and compute the gradient $\partial F/\partial C_t$. 
We accumulate these gradients along the interpolation path.
We keep the generated sequence ${\mathcal{T}}_\text{out}$ fixed and define the scalar policy score as
\begin{equation}
    F(\widehat{E}_t^n,\mathcal{T}_\text{out}) = \frac{1}{n_t} \sum_{j=1}^{n_t} \log p_\theta \left( {y}_j \mid \widehat{E}_t^n \right). 
\end{equation}
Thus, $F$ measures how strongly the model supports the action sequence $\mathcal{T}_\text{out}$. 
The attribution for input token  $i$ is approximated using the $N$ interpolated observations 
\begin{equation}
S_{\mathrm{IG}}(i) \approx (e_{t,i}-\widetilde{e}_{t,i}) \odot \frac{1}{N} \sum_{n=1}^{N} \frac{ \partial F\left( \widehat{E}_t^{\,n}; \mathcal{T}_{A_t} \right) }{ \partial {e}_{t,i} },
\end{equation}
where ${e}_{t,i}$ is the embedding of token $i$ at time $t$.

\subsubsection{Token Activation Map (TAM)~\cite{li2025tam}}
TAM measures the logit value of each input token in the generated token. Consider $H_t\in\mathbb{R}^{n_t\times d}$ the hidden representations of the tokens in $C_t$, and let $w_v\in\mathbb{R}^d$ denote the language-model-head vector associated with vocabulary token $v$. For generated token $y_j$, its activation over the input is $\Gamma_j=\operatorname{ReLU}(H_t w_{y_j})$.
To account for autoregressive context, TAM removes the activation explained by preceding tokens, yielding a refined map $\overline{\Gamma}_j$ as in~\cite{li2025tam}. The final token score averages these refined activations over the generated action tokens as
\begin{equation}
S_{\text{TAM}}(i)=\frac{1}{n_t}\sum_{j=1}^{n_t}\overline{\Gamma}_{j,i}.
\end{equation}

\subsection{Steering}
Input interventions characterize which mode affects a policy's decision.
Steering complements this analysis through internal interventions, all while keeping model parameters fixed.
We study two steering methods:

\subsubsection{Value-Vector-Guided Semantic Steering~\cite{geva2022transformer}}
We choose the feed-forward network (FFN) of a transformer block of the VLM backbone. Let $z_{l}\in\mathbb{R}^{m_{l}}$ denote the input activation to the FFN's down projection at layer $l$, and $m_l$ the intermediate dimension in the FFN.
Let
$W_l\in\mathbb{R}^{d\times m_l}$ be the down-projection weight where $d$ denotes the transformer hidden state's dimension. Its output is
\begin{equation}
\operatorname{FFN}_{l}
= W_{l} z_{l}
= \sum_{j=1}^{m_l}z_{l,j}v_{l,j}\;;\;
v_{l,j}=W_l[:,j].
\end{equation}
Following the value-vector interpretation in
\cite{haon2025mechanistic,geva2022transformer}, we associate each fixed value vector with its highest-scoring vocabulary tokens. 
For a concept token set $\mathcal{T}(c)$, we select
\begin{equation}
\mathcal{S}_l(c)=\left\{j:
\operatorname{TopK}({W}_{LM}v_{l,j})\cap\mathcal{T}(c)\ne\varnothing
\right\},
\end{equation}
where $W_{LM}$ is the vocabulary output projection, i.e., LM head. 
At inference, we replace $z_{l,j}$ by a prescribed scalar $\beta$ for selected neurons and leave other activations unchanged. 
This changes the contribution of the selected value vectors without modifying their weights. We also choose semantically similar positive clusters $ S_l^{+}$ and semantically opposite negative clusters $S_l^{-}$, and clamp them accordingly. 
The change in the FFN output $\Delta$ after steering is
\begin{equation}
\Delta
=
\sum_{j \in S_{l}^{+}}
\left(\beta_{+} - z_{l,j}\right) v_{l,j}
-
\sum_{j \in S_{l}^{-}}
\left(\beta_{-} + z_{l,j}\right) v_{l,j}.
\end{equation}

\subsubsection{Activation Steering~\cite{ActAddturner2023steering}}
We also construct behavior-associated directions from recorded hidden states.
Let $\mathcal{D}_{+}$ and $\mathcal{D}_{-}$ contain activations associated with positive and negative concepts. 
At the specified prediction position, their layer-wise mean difference is
\begin{equation}
 d_l = \frac{1}{|\mathcal{D}_{+}|}
 \sum_{\mathcal{D}_{+}}h_{l_+}
 -\frac{1}{|\mathcal{D}_{-}|}
 \sum_{\mathcal{D}_{-}}h_{l_-}.
\end{equation}
where $h_l$ is the output hidden state at some layer $l$.
For selected layers and token positions, we apply
\begin{equation}
 \widetilde h_l=h_l+\alpha d_l,
\end{equation}
where $\alpha$ controls intervention strength to steer the model from negative to positive behaviors.
\section{Experiments}

\subsection{VLN Baselines}

For this study, we compare CorrectNav~\cite{yu2026correctnav}, DualVLN~\cite{wei2026ground}, AwareVLN~\cite{guo2026awarevln}, and NaVILA~\cite{cheng2024navila}, which are VLM-based.
Although all four methods condition on a language instruction and egocentric RGB observations, they differ in their backbones, visual memory sampling, reasoning mechanisms, action spaces, camera configurations, and prompt formats. 
CorrectNav, DualVLN, and AwareVLN report a similar success rate of $\sim$65\% in the VLN-CE R2R benchmark~\cite{R2R,krantz2020VLN-CE}, whereas NaVILA reports $\sim$54\%.
We run all of our experiments on a subset of validation unseen in R2R VLN-CE. 
We ran each baseline in its nominal camera, forward, and turning step configurations, which thus differ between methods.

\subsection{Sensitivity to Visual Semantics\label{sec:iss-img}}

\begin{figure*}[t]
    \centering
    \vspace{5.1pt}
    \begin{minipage}[t]{0.70\textwidth}
        \centering
        \includegraphics[width=\linewidth]{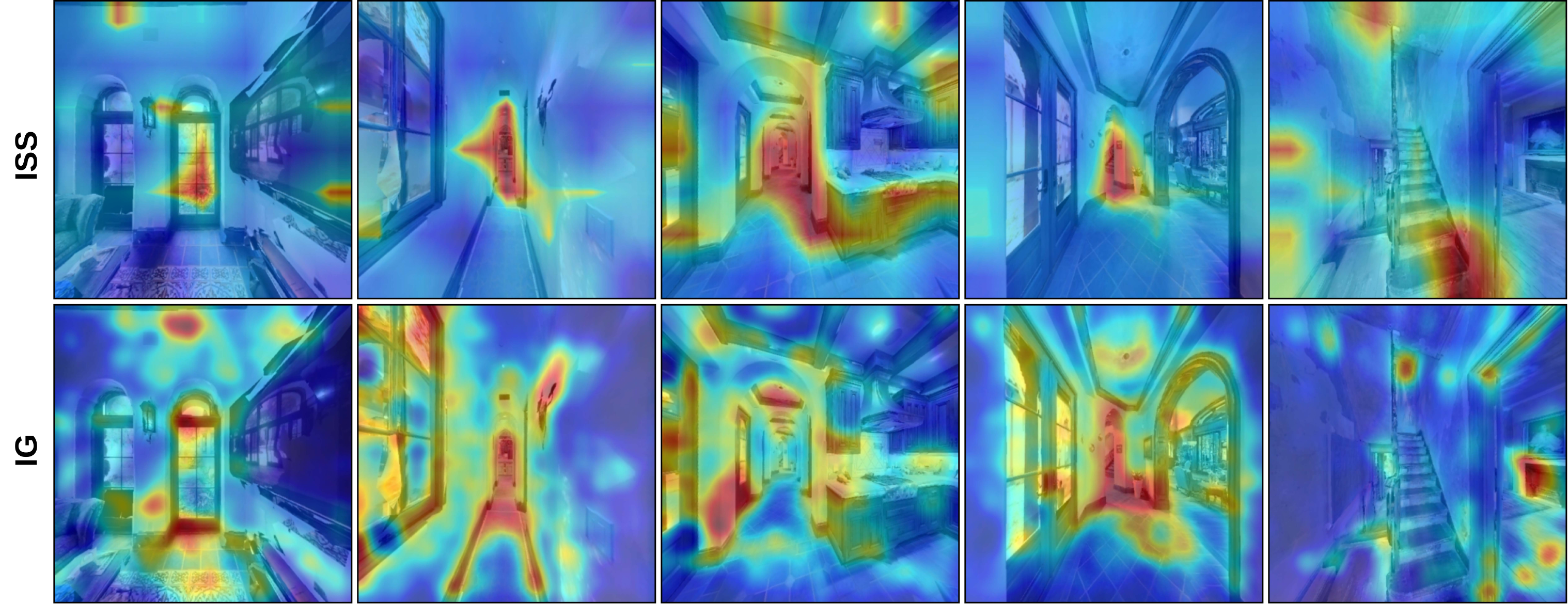}

        {\footnotesize{(a)} Heatmaps using Interventional Significance Score (ISS) and Integrated Gradients (IG).}
    \end{minipage}
    \hfill
    \begin{minipage}[t]{0.27\textwidth}
        \centering
        \includegraphics[width=\linewidth]{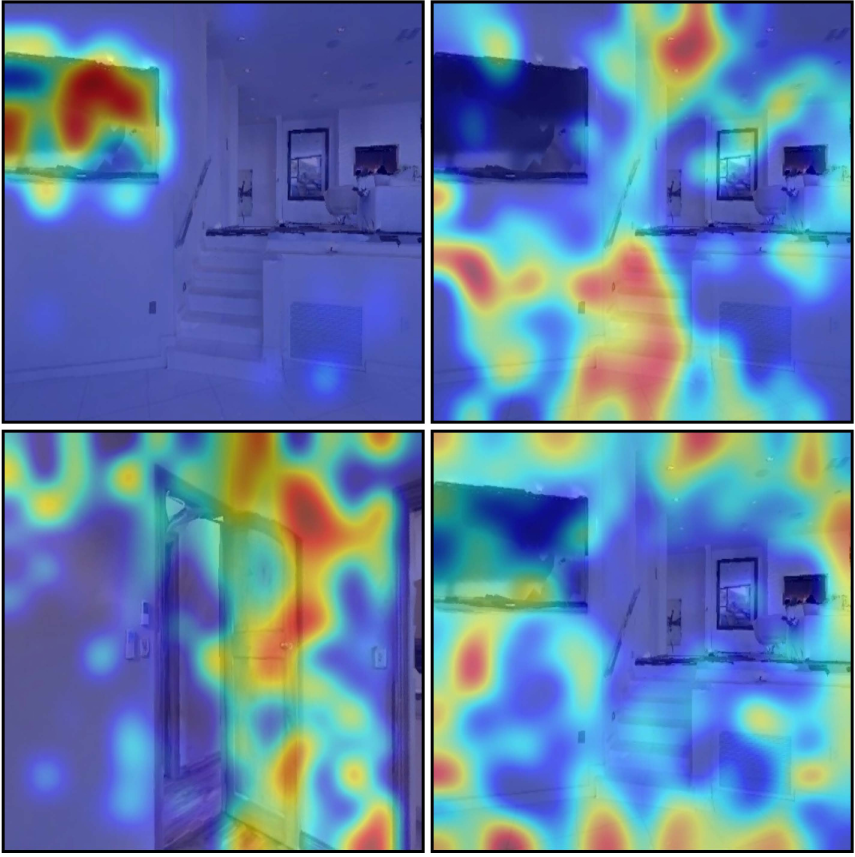}

        {\footnotesize{(b)} Token Activation Map (TAM).}
    \end{minipage}

    \caption{
        Qualitative analysis of the heat maps from ISS vs. IG (left) and TAM (right).
        (a) ISS and IG heatmaps relate, from left to right, to the subtasks:
        \textit{stop by the window}, \textit{go down the hall} , \textit{walk through the left of the stove}, \textit{walk towards the stairs}, and \textit{go up the stairs}.
        ISS heatmaps are more concentrated on the relevant objects, whereas IG heatmaps are more diffused.
        (b) TAM heatmaps correspond to the tokens \textit{screen} (top left), \textit{stairs} (top right), \textit{door} (bottom left), and \textit{move forward} (bottom right). 
        TAM localizes semantic objects effectively, but produces noisier activations for abstract navigation actions predicted by VLN policies.
    }
    \label{fig:qualitative}

    \refstepcounter{subfigref}\label{fig:iss_ig}
    \refstepcounter{subfigref}\label{fig:tam}
\end{figure*}

We evaluate ISS and IG explanations on \(200\) open-loop decision points sampled from the dataset. 
For each point, we manually annotate the image regions that are visually important for selecting the next navigation action. 
We use these annotations as a common evaluation reference across all models and explanation methods.

Given a generated explanation heatmap $S$ and its corresponding manual annotation mask $M$, we compute Top-$k$ Precision over the most salient pixels of the heatmap. 
Let $\Omega_k(S)$ denote the set of pixels whose saliency values lie in the top $k\%$ of $S$. 
The precision score is defined as
\begin{equation}
    \mathrm{Prec@}k = \frac{\sum_{(u,v) \ \in \ \Omega_k(S)} M_{u,v}}{|\Omega_k(S)|}.
\end{equation}
This metric measures what fraction of the high-attribution region lies inside the manually annotated area.

We emphasize that our goal is not to perform semantic segmentation,
rather, the annotations provide a human-defined reference for visual evidence important at a particular decision point. 
Therefore, we focus on precision rather than recall, F1, or mean IoU, where high precision identifies whether the $k$\% peaks of the explanation map are concentrated on the annotated decision-relevant regions.

The precision metrics are shown in Table~\ref{tab:modelwise_iss_ig_precision}.
Across our experiments, ISS explanations obtain stronger annotation alignment than IG explanations.
This might be because gradients struggle to propagate from the output likelihood through the entire model back to the input pixels (i.e., vanishing gradients). 
This long gradient path produces diffuse attribution maps, as shown in Fig.~\ref{fig:iss_ig}. 
In contrast, ISS measures behavioral sensitivity by perturbing the visual input and observing changes in the model's navigation behavior. 

We observe that the differences between models are most apparent at lower $k$ values. 
For example, CorrectNav assigns higher attribution to the annotated regions at low $k$, indicating stronger peak visual focus. 
In contrast, NaVILA shows weaker alignment at low $k$, suggesting that its most salient ISS regions are less concentrated on the annotated areas. 
As $k$ increases, the metric includes broader portions of the heatmap, and the gap between models becomes less focused on peak localization and more reflective of wider attribution coverage.
In conclusion, the ISS and IG precisions of $\sim$40-60\% across all methods suggest that, while a decent share of the visual attention lies on relevant cues, current VLN pipelines are roughly equally sensitive to cues deemed irrelevant by a human. This sensitivity might be a residual of spurious correlations during training or indicative of limited training data diversity. 

Additionally, we perform a qualitative analysis over the TAM heatmaps. 
As seen in Fig.~\ref{fig:tam}, TAM produces relatively interpretable localization for semantically grounded concepts with clear visual references.
However, its behavior is considerably less reliable for abstract VLN output tokens such as \textit{move forward}. 
Since such tokens do not correspond to a single, well-defined visual entity in the scene, the resulting activation maps tend to be diffuse and difficult to interpret. 
This suggests that TAM is better suited to visualizing object- or concept-level semantics than explaining abstract navigation decisions.

\begin{table}[t]
    \vspace{5.1pt}
\centering
\caption{
Top-\(k\) Precision comparison between ISS and IG.
}
\label{tab:modelwise_iss_ig_precision}
\small
\fontsize{7.5}{9}\selectfont
\renewcommand{\arraystretch}{1.2}
\begin{tabular}{l|l|c|c|c|c}
\noalign{\hrule height 1pt}
Model & Method & P@1 & P@10 & P@30 & P@50 \\
\noalign{\hrule height 1pt}
\multirow{2}{*}{CorrectNav}
& ISS & \textbf{\underline{0.66}} & \underline{0.52} & \underline{0.41} & 0.36 \\
& IG  & 0.52 & 0.46 & 0.4 & \underline{0.37} \\
\hline
\multirow{2}{*}{AwareVLN}
& ISS & \underline{0.60} & \textbf{\underline{0.55}} & \textbf{\underline{0.44}} & \textbf{\underline{0.40}} \\
& IG  & 0.46 & 0.43 & 0.40 & 0.38 \\
\hline
\multirow{2}{*}{DualVLN}
& ISS & \underline{0.55} & \underline{0.49} & \underline{0.43} & 0.39 \\
& IG  & 0.42 & 0.42 & 0.41 & \underline{0.40} \\
\hline
\multirow{2}{*}{NaVILA}
& ISS & \underline{0.48} & 0.42 & 0.37 & 0.34 \\
& IG  & 0.47 & \underline{0.44} & \underline{0.40} & \underline{0.37} \\
\noalign{\hrule height 1pt}
\end{tabular}
\begin{center}
    \footnotesize 
    Highest value across policies in bold; highest value per policy underlined.
\end{center}
    \vspace{-1em}
\end{table}

\subsection{Sensitivity to Visual Memory\label{sec:mem-iss-img}}

To evaluate how past observations influence navigation decisions, we compute ISS by blurring a percentage $p$ of the images in memory, from oldest to most recent.
We run perturbations for 12 episodes with three variations of the verbal command per trajectory in all scenes of the validation unseen split.
For a fair comparison between pipelines, we run each baseline over the ground-truth trajectory and compute the ISS score.
Note that the trajectory horizon of AwareVLN and NaVILA is 1 to 3 steps, whereas CorrectNav and DualVLN execute 3 to 4 steps.
Thus, one cannot directly compare ISS between the two horizon groups.
We instead report normalized ISS scores per method to provide insights about their sensitivity patterns.

\begin{figure}[t]

    \vspace{5.1pt}
    \centering
    \includegraphics[width=\linewidth]{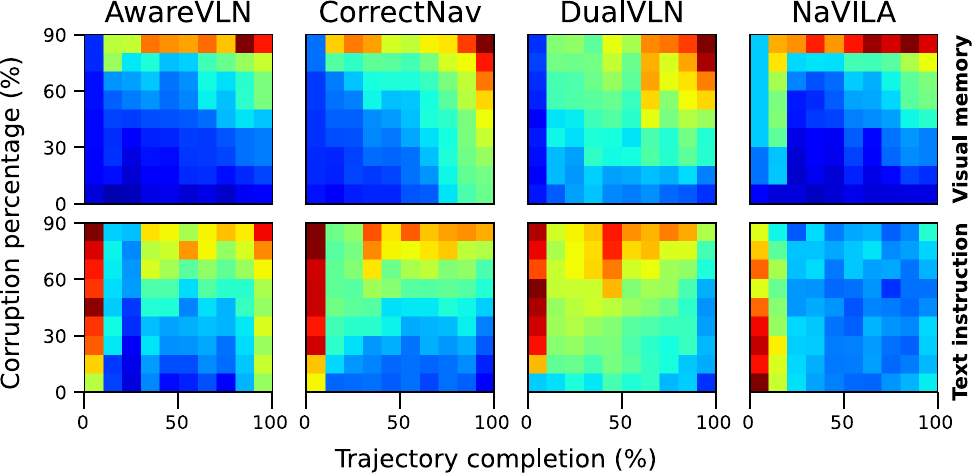}
    \caption{ISS heatmap of visual-memory (top) and textual memory (bottom) intervention across trajectory progress. 
    Each cell reports the mean ISS between the unperturbed action and the action predicted after blurring a given fraction of the images currently stored in memory or substituting part of the navigation instruction with its antonyms.
    All baselines are compared on the same ground truth trajectories.
    ISS scores increase from blue to red.
    }
    \label{fig:blur-iss}
\end{figure}

Fig.~\ref{fig:blur-iss} shows the mean ISS score per trajectory length and memory corruption percentage in accumulated bins.
A natural expectation is that ISS increases with trajectory progress, since early memory observations are often highly correlated, whereas later ones span a longer history and provide more distinct cues about the agent’s progress. 
Note that monotonic vertical growth in the ISS is not required, since ISS may not change if historical images become redundant or localized if only particular landmarks matter for the current decision.
CorrectNav and DualVLN show increased sensitivity with both trajectory progress and corruption strength, indicating stronger dependency on information distributed across their visual history. 
DualVLN exhibits sensitivity to visual-memory degradation earlier in the trajectory than CorrectNav.
This may stem from DualVLN's hierarchical structure: historical observations are distilled into a latent bottleneck before passing through the diffusion policy, whereas CorrectNav outputs actions directly from the VLM, potentially allowing the instruction to compensate for corrupted historical cues.
CorrectNav also explicitly annotates temporal tags for the history frames in the prompt, potentially encouraging the use of distributed memory cues at later timesteps.

AwareVLN and NaVILA show weaker accumulation effects, indicating a stronger reliance on a more recent subset of observations.
Although NaVILA and AwareVLN share the same memory sampling schema, they differ in their reasoning paradigm.
AwareVLN uses chain-of-thought reasoning that can reaffirm or correct perceived progress, adding redundant information to the prompt that can compensate for corrupted visual memory.
This might explain why AwareVLN's overall sensitivity to memory corruption is lower than NaVILA's, which has no other source of progress tracking than the visual memory.
NaVILA shows increased sensitivity during the earliest timesteps, reflecting a cold-start regime of the memory where initial observations may disproportionately matter for understanding orientation (i.e., away from a wall). 

Overall, the heatmaps show a similar pattern where reliance on visual memory tends to increase as the trajectory progresses, and larger parts of the memory become relevant as more distinct observations are gathered and longer context becomes necessary for progress tracking.
The degree of this effect varies between the different VLN architectures.
Yet this dependence is not exclusive; complementary cues from language, current observations, and internal state may be compensating for degraded memory and preserving the action prediction.

\subsection{Robustness to Visual Memory}
It is important to point out that sensitivity analyses how strongly perturbing past visual observations changes the action predicted at a fixed state along the ground-truth trajectory. 
It does not indicate whether the original or perturbed action is correct during closed-loop execution. 
To evaluate \emph{robustness} to closed-loop memory degradation, Table~\ref{tab:sr_blur_levels} reports the success rate versus the fraction of early visual memory that is blurred.
At low and moderate corruption levels, all pipelines remain relatively robust, with only limited degradation up to about \(50\%\) memory corruption, after which their behaviors diverge.
Despite exhibiting substantial action sensitivity to memory blurring, CorrectNav has the most robust closed-loop performance when only the current image is available. 
Similarly, while AwareVLN has lower open-loop sensitivity, it suffers larger closed-loop success-rate degradation than the other VLNs.
This corroborates that the consequences of memory corruption notably depend on the policy's ability to tolerate or recover from altered decisions using the other input modes.

\begin{table}
\centering
\caption{
Success Rate at different image memory corruption percentages on a subset of R2R val-unseen. 
}
\label{tab:sr_blur_levels}
\small
\renewcommand{\arraystretch}{1.2}
\resizebox{\columnwidth}{!}{
\begin{tabular}{l|c|c|c|c|c|c|c|c|c|c}
\noalign{\hrule height 1pt}
Model & 0\% & 10\% & 20\% & 30\% & 40\% & 50\% & 60\% & 70\% & 80\% & 90\% \\
\noalign{\hrule height 1pt}
CorrectNav
& 55.97 & 54.48 & 52.99 & 48.51 & \textbf{51.49}
& \textbf{50.00} & \textbf{47.01} & \textbf{47.76} & \textbf{44.03} & \textbf{35.82} \\
DualVLN
& 48.51 & 47.76 & 47.76 & 42.54 & 50.75
& 44.03 & 33.58 & 30.60 & 32.84 & 21.64 \\
AwareVLN
& \textbf{58.21} & \textbf{58.21} & \textbf{57.46} & \textbf{52.24} & 48.51
& 49.25 & 35.82 & 32.84 & 32.09 & 15.67 \\
NaVILA
& 50.00 & 46.27 & 48.51 & 42.54 & 41.79
& 48.51 & 37.31 & 32.84 & 23.13 & 14.18 \\
\noalign{\hrule height 1pt}
\end{tabular}
}
\end{table}

\subsection{Sensitivity to Textual Semantics\label{sec:iss-text}}

We evaluate ISS on 100 open-loop samples where we corrupt only the navigation instructions by deleting groups of 1, 3, or 5 words, leaving the rest of the prompt unchanged.
For each sample, we manually annotated which semantic concepts in the language instruction are visually important for the next navigation decision.
Since human task-following typically relies on a set of semantic cues rather than a single isolated concept, our masks are full subtasks, e.g., ``go down the stairs and turn right'' rather than ``stairs'' or ``right'' only.

\begin{table}[t]
    \vspace{5.1pt}
\centering
\caption{
Top-\(k\) precision (\%) and percentages of trajectories with zero ISS scores (ZS) for text corruption.
}
\label{tab:modelwise_topk_precision_text}
\small
\fontsize{7.5}{9}\selectfont
\renewcommand{\arraystretch}{1.2}
\begin{tabular}{l|c|c|c|c|c|c}
\noalign{\hrule height 1pt}
Model & P@1 & P@3 & P@5 &
P@10 & P@15 & ZS(\%) \\
\noalign{\hrule height 1pt}
CorrectNav
& \textbf{50.93} & \textbf{51.54} & \textbf{50.00}
& \textbf{44.84} & \textbf{37.95} & 32.41 \\
AwareVLN
& 41.67 & 43.21 & 42.96 & 39.35 & 36.06 & 48.15 \\
DualVLN
& 40.74 & 43.83 & 42.04 & 36.84 & 34.05 & 20.37 \\
NaVILA
& 34.26 & 32.10 & 32.41 & 29.58 & 29.45 & 2.78 \\
\noalign{\hrule height 1pt}
\end{tabular}
\end{table}

Fig.~\ref{fig:iss_text} shows qualitative results of the ISS computed over the textual instruction.
Active subtask verbs and semantic cues show higher sensitivity, even when farther from the camera.
Table~\ref{tab:modelwise_topk_precision_text} reports precision results. CorrectNav consistently achieves the highest precision across all $k$, indicating that the words producing the largest ISS changes are within the human-annotated decision-relevant concepts. 
In contrast, AwareVLN has the most zero-ISS samples, meaning the baseline trajectory is robust to corruption and can infer the deleted segment from nearby context.
NaVILA combines the lowest precision with almost no zero-score cases, suggesting high dependence on the current active subtask text.
Many low-precision cases arise when the model is sensitive to semantically related cues outside the manually annotated span. 
These often correspond to repeated concepts that remain strongly grounded in either visual memory or the current subtask.
Overall, sensitivity to text semantic cues is brittle and therefore not a strong indicator of input focus, since landmarks can repeat across the entire history.

\subsection{Sensitivity to Textual Memory}

We evaluate how textual memory context influences navigation decisions.
To ensure that each intervention produces a meaningful perturbation, we replace a percentage of the navigation instruction with antonyms rather than deleting them. 
This reduces the chance of the model recovering the missing information from the surrounding context and introduces semantically conflicting cues to induce deviations from the unperturbed action.
Because antonym substitutions are introduced cumulatively from the beginning of the instruction and the experiments run over the ground-truth trajectory, low corruption affects early timesteps, while stronger corruption progressively reaches instruction content relevant to later decisions. 
The boundary between low and high ISS should thus shift toward the right with trajectory completion.

Fig.~\ref{fig:blur-iss} shows the mean ISS score per trajectory length and textual memory corruption percentage in accumulated bins.  
CorrectNav and DualVLN show that later decisions remain stable until corruption reaches farther into the instruction, suggesting that the portion of language influencing the action shifts as navigation progresses.
DualVLN is affected earlier in the trajectory, which may again originate from its hierarchical design.
The first trajectory columns are consistently red because early decisions depend most on the beginning of the instruction, which is corrupted first.

AwareVLN and NaVILA use similar visual memory sampling, but show notably different sensitivities. 
At earlier subtasks, AwareVLN depends less on the text than at the end.
Architecturally, AwareVLN’s recurrent reasoning state explicitly summarizes task progress, creating a competition between an internal belief that the task is complete and increasingly corrupted linguistic cues.
It remains unclear why AwareVLN behaves unstably at the end of the trajectory, independent of the corruption level.

NaVILA's sensitivity remains concentrated early in the trajectory, while later decisions remain comparatively insensitive.
This suggests that the instruction strongly affects the initial route decision, but subsequent actions are primarily driven by the visual observations. 
This does not mean NaVILA ignores later instruction content, but the heatmap provides weaker evidence that its linguistic sensitivity is synchronized with trajectory progress.
In summary, the heatmaps show no apparent pattern across all VLNs.

\subsection{Semantic Concepts from VLMs in VLNs}

To assess whether the original semantic concepts of the pre-trained VLM remain in the fine-tuned VLNs, we compare the value vectors of the original Qwen2.5-VL model and DualVLN. 
DualVLN is trained to output symbolic text rather than the textual tokens that CorrectNav and AwareVLN are trained on.
For each value vector $v_{l,j}$, we identify its top 20 tokens.
We then analyze the tokens \textit{forward}, \textit{left}, \textit{right}, \textit{stop}, and the arrow symbols, and count how many value vectors per layer correspond to these tokens.

\begin{figure}
    \vspace{5.1pt}
    \centering
    \includegraphics[width=\linewidth]{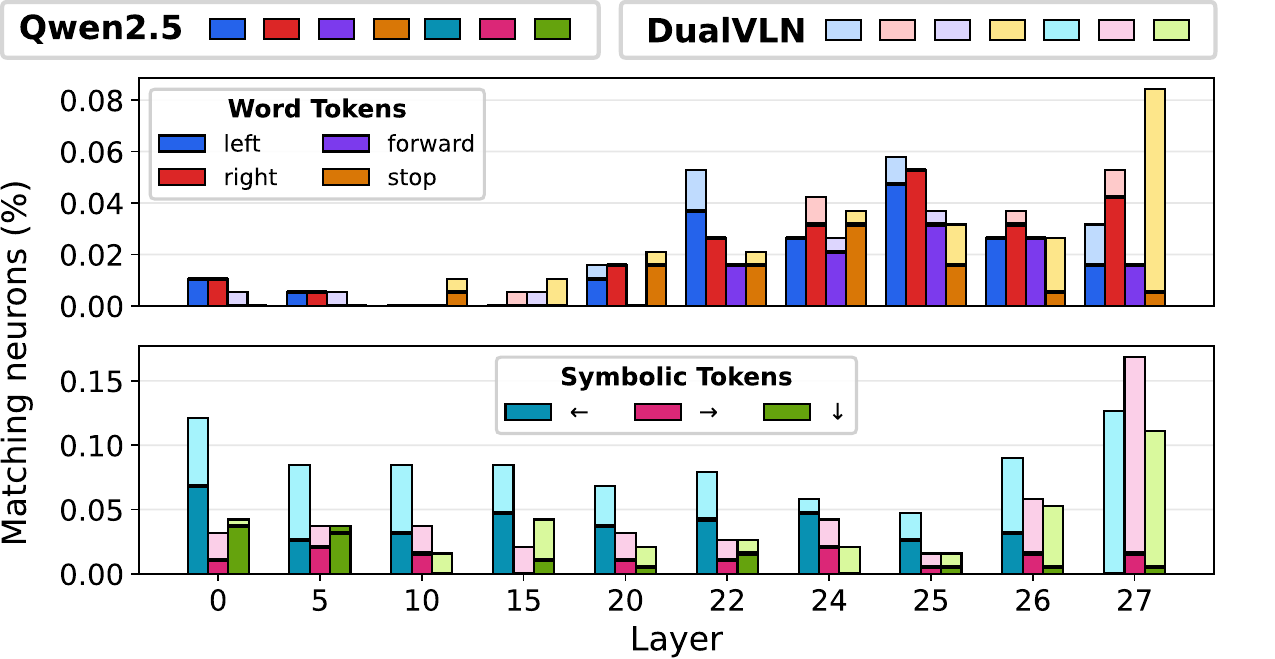}
    \caption{Percentage of value vectors for DualVLN (lighter) and Qwen2.5-VL (darker) model for textual and symbolic navigation tokens (top 20 tokens).
    Each layer has 18,944 neurons.
    VLN training does not lose the concepts already present in the original VLM. 
    Fine-tuning introduces activations for symbolic tokens, particularly in the last layers.
    The thick black line shows the shared overlap height between bars of different models.
    To avoid cluttering, we do not plot all layers of the model.
    }
    \label{fig:sem-holding}
\end{figure}

Fig.~\ref{fig:sem-holding} shows that activation of the semantic navigation direction textual tokens are not lost with fine-tuning, even though DualVLN is trained to generate symbolic tokens.
Only in the 5th layer the activation of a symbolic token is smaller for DualVLN than for the original Qwen2.5 model.
DualVLN outputs a verbal \textit{stop}, hence the increased activation at the last layer, similar to the symbolic tokens.

\begin{table}[t]
\centering
\fontsize{7.5}{9}\selectfont
\caption{Change in motion tokens with neuron clamping.
\label{tab:steering_studies}}
\renewcommand{\arraystretch}{1.2}
\begin{tabular}{l|c|c|c}
\noalign{\hrule height 1pt}
\textbf{Steering / Motion} & Forward & Right & Left \\
\hline
{\textit{forward} + $\,\mid\,$ \textit{turn} -} & +3.2\% & -0.8\% & -19.3\% \\
{\textit{left} + $\,\mid\,$ \textit{right} -} &  +0.8\% & -3.0\% & +7.8\% \\
\noalign{\hrule height 1pt}
\end{tabular}
\end{table}

To show that the semantic concepts still survive VLN fine-tuning and can causally influence navigation decisions, we conduct activation-clamping experiments.
We fix selected neurons and set their input activations to predefined values if they are associated with token $k$, i.e., $k$ appears among its top-20 vocabulary activations.
We run two clamping experiments over 5 episodes, contrasting \textit{forward} with \textit{turn} and \textit{right} with \textit{left}. 
We clamp the positive concept $S_{+}$ to $\beta_{+}=10$ and the negative $S_{-}$ to $\beta_{-}=10$.
Table~\ref{tab:steering_studies} shows that this biases navigation toward the corresponding semantic behaviors, even though these tokens are not part of DualVLN's output space and the actions are generated by a diffusion policy.
These results show that semantic language concepts retained from the original VLM, in spite of VLN training, could provide a solution to steering VLN behaviors.

\subsection{Activation Reasoning Under Visual Distribution Shift}
\begin{figure}
    \vspace{5.1pt}
    \centering
    \includegraphics[width=\linewidth]{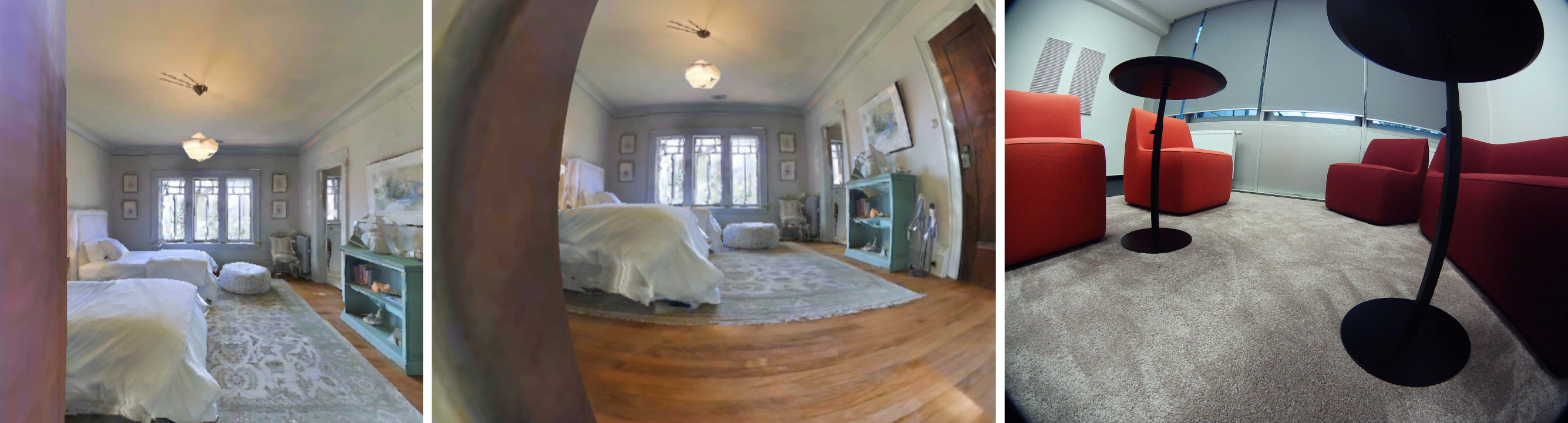}
    \caption{ (Left) The pinhole camera the baseline is trained on (Center) the visual distribution shift in simulation
    (Right) real robot configuration we use for experiments 
    }
    \label{fig:real-world-deployment}
\end{figure}

We next examine whether an existing explicit reasoning behavior remains
accessible when its occurrence decreases under a distribution shift.
Unlike semantic steering, which targets motion preferences, this experiment
targets AwareVLN's choice between explicit reasoning and direct action
prediction.
We compare nominal-camera and fisheye observations at a lower height with matched simulator states, retaining the same navigation instructions and trajectory history.
The changed camera is a fisheye with a $\xi=$-0.20 and an $\alpha=$0.59 with a focal length of 218.93 for both axes~\cite{usenko2018double}. 
We choose these parameters to match the camera used for robot deployment, as shown in Fig.~\ref{fig:real-world-deployment}. 

We measure the fraction of inference calls that emit explicit reasoning in simulation.
As shown in Table~\ref{tab:reasoning_frequency}, the reasoning rate drops 4$\times$ under the visual distribution shift.  
To remedy this drop, we use activation steering. 
We choose one target layer and add the mean activation vector to shift the concept from action to reasoning in AwareVLN's first output token, pushing the model to reason.
We evaluate steering strength on 306 decision contexts under the same camera parameters as before.
Note that the reasoning frequency increases with $\alpha$, saturating at $\alpha=4$.
We also see that random-vector steering does not improve this frequency, showing the effectiveness of the designed mean vector.

\subsection{Real World Deployment of Activation Steering}
We demonstrate the generalization of the identified steering vectors to the real world by evaluating over 40 trials across 8 scenes and 12 instructions using a sequential recovery protocol. 
We consider the policy to have failed if it commits to the wrong choice and does not correct itself, or if it gets stuck in a loop. 
The unmodified ($\alpha=0$) policy runs first. 
When an operator identifies a failure based on the criterion above, we apply an activation intervention with $\alpha=3$ in the target layer on the first token, forcing the policy to recover using reasoning. 
We emphasize that we compute the mean vector for the action-to-reasoning data in simulation, which uses the pinhole camera. 
This allows us to induce reasoning without changing parameters or collecting additional data in the new modality, directly in the real world.  

AwareVLN's non-steered variant had a $37.2\%$ success rate, whereas our steered variant succeeded in $74.4\%$ of the trials.
This represents a $59.3\%$ recovery from failure states, without fine-tuning the model with out-of-distribution data.
These results show that the identified steering can be translated zero-shot to the real world and that the induced interventions can significantly improve the success rate.
While our studies were causal and required intervention after recognizing failure, more research is needed to develop introspective methods that can trigger steering online.

\begin{table}[t]

    \vspace{5.1pt}
\centering
\caption{Reasoning frequency under visual distribution shift and steering interventions.}
\label{tab:reasoning_frequency}
\fontsize{7.5}{9}\selectfont
\renewcommand{\arraystretch}{1.2}
\begin{tabular}{l|c|c|c|c}
\noalign{\hrule height 1pt}
\textbf{Condition} & \multicolumn{4}{c}{\textbf{Reasoning Frequency}} \\
\noalign{\hrule height 1pt}
Baseline Visual Distribution
& \multicolumn{4}{c}{$12 \pm 2\%$} \\

Visual distribution shift
& \multicolumn{4}{c}{$3 \pm 1\%$} \\

\hline

${\alpha}$
& ${0.5}$
& ${1}$
& ${2}$
& ${4}$ \\

\hline

Random vector steering
& $3.0\%$
& $2.7\%$
& $2.6\%$
& $1.6\%$ \\

Mean vector steering
& ${20\%}$
& ${55\%}$
& ${90\%}$
& ${100\%}$ \\

\noalign{\hrule height 1pt}
\end{tabular}
\end{table}

\section{Conclusions}

This work presents the first causal interpretability study for VLM-based VLNs.
Our results suggest that while VLN policies are sensitive to many input and memory modalities, there is no apparent correlation between sensitivity and success rate.
Robustness appears to be primarily governed by a VLN policy's ability to recover and adapt using other input modalities.
Surprisingly, our sensitivity studies to semantic concepts show that while there is some overlap with concepts relevant to humans, there is still similar sensitivity to concepts and areas deemed irrelevant. 
This might be because the models are overfitting, the data is not sufficiently diverse, or the models rely on shortcuts.
We also show that semantic concepts from the original VLM survive and can be steered in the fine-tuned VLN, even if they are not part of its output action space.
Finally, we demonstrate zero-shot activation steering to induce model reasoning in real-world visual-distribution shift. 
We show that this intervention, learned purely in simulation, can significantly improve success rate in out-of-domain scenarios without changing the network weights.
Future work could study whether a pattern emerges in the lack of focus on semantic cues, for instance, in navigable paths.
Another possible branch is using controllers for online tuning of $\alpha$ and operator-free activation steering.

\section*{Acknowledgments}
\blackout{
 The authors gratefully acknowledge the scientific support and HPC resources provided by the Erlangen National High Performance Computing Center (NHR@FAU) of the Friedrich-Alexander-Universität Erlangen-Nürnberg (FAU) under the BayernKI project v106be. BayernKI funding is provided by Bavarian state authorities. The authors also acknowledge the Gauss Centre for Supercomputing e.V. for funding this project by providing computing time on the GCS Supercomputer JUPITER at Jülich Supercomputing Centre. We also acknowledge the support by the Körber European Science Prize. This work has been partially supported by the German Federal Ministry of Research, Technology and Space (BMFTR) under the Robotics Institute Germany (RIG). 
}

\bibliographystyle{IEEEtran}
\bibliography{root}

\end{document}